\documentclass[journal,twoside,web]{ieeecolor_arxiv}

\usepackage{tmi}
\usepackage{cite}
\usepackage{amsmath,amssymb,amsfonts}
\usepackage{algorithmic}
\usepackage{graphicx}
\usepackage{textcomp}
\usepackage{multirow}
\usepackage{float}
\usepackage{amsmath}
\usepackage{bbding}
\usepackage{pifont}

\usepackage{threeparttable} 

\usepackage{booktabs} 
\usepackage{subcaption} 
\usepackage{amsfonts}
\usepackage[misc]{ifsym}
\usepackage{url}
\usepackage{hyperref}

\newcommand{\rraw}{\mathbf{I}^{raw}}
\newcommand{\ra}{\mathbf{I}^a}
\newcommand{\rp}{\mathbf{I}^p}
\newcommand{\ea}{\mathbf{T}^a}
\newcommand{\ep}{\mathbf{T}^p}
\newcommand{\eai}{\mathbf{T}^{a_i}}
\newcommand{\epj}{\mathbf{T}^{p_j}}
\newcommand{\raset}{\{\mathbf{I}^a_q\}_n}
\newcommand{\rpset}{\{\mathbf{I}^p_q\}_m}

\newcommand{\rpq}{\mathbf{I}_q^p}
\newcommand{\zaa}{\mathbf{Z}_1^a}
\newcommand{\zab}{\mathbf{Z}_2^a}
\newcommand{\zpa}{\mathbf{Z}_1^p}

\newcommand{\easet}{\{\mathbf{T}^a\}_n}
\newcommand{\epset}{\{\mathbf{T}^p\}_m}

\newcommand{\qa}{Q^a}
\newcommand{\qp}{Q^p}

\newcommand{\Lap}{\mathbf{y}^{a,p}}
\newcommand{\lap}{y^{a,p}}

\newcommand{\Lp}{\mathbf{y}^p}
\newcommand{\Licl}{\mathcal{L}_{ICL}}
\newcommand{\Lproto}{\mathcal{L}_{ProtoCL}}
\newcommand{\Lexist}{\mathcal{L}_{Exist}}
\newcommand{\Sap}{\mathbf{S}^{a,p}}
\newcommand{\yhat}{\hat{y}}
\newcommand{\yhatmatrix}{\mathbf{\yhat}}
\newcommand{\ourmodel}{MeDSLIP}

\def\BibTeX{{\rm B\kern-.05em{\sc i\kern-.025em b}\kern-.08em
    T\kern-.1667em\lower.7ex\hbox{E}\kern-.125emX}}
\begin{document}
\title{\ourmodel: Medical Dual-Stream Language-Image Pre-training with Pathology-Anatomy Semantic Alignment}
\author{Wenrui Fan, Mohammod N.I. Suvon, Shuo Zhou, Xianyuan Liu, Samer Alabed, Venet Osmani, Andrew J. Swift, Chen Chen, and
    Haiping Lu, \IEEEmembership{Senior Member, IEEE}
\thanks{This work has been submitted to the IEEE for possible publication. Copyright may be transferred without notice, after which this version may no longer be accessible.}
\thanks{Wenrui Fan, Mohammod N.I. Suvon, Shuo Zhou, Xianyuan Liu, and Haiping Lu are with Centre for Machine Intelligence and School of Computer Science, University of Sheffield, S1 4DP Sheffield, U.K. (e-mail: wenrui.fan@sheffield.ac.uk; m.suvon@sheffield.ac.uk; shuo.zhou@sheffield.ac.uk; xianyuan.liu@sheffield.ac.uk; h.lu@sheffield.ac.uk) (Corresponding author: Haiping Lu).}
\thanks{Samer Alabed and Andrew J. Swift are with School of Medicine and Population Health, and INSIGNEO, Institute for in Silico Medicine, University of Sheffield, S10 2TN Sheffield, and Department of Clinical Radiology, Sheffield Teaching Hospitals, S10 2JF Sheffield, U.K. (e-mail: s.alabed@sheffield.ac.uk; a.j.swift@sheffield.ac.uk).}
\thanks{Venet Osmani is with Digital Environment Research Institute, Queen Mary University of London, E1 1HH London, U.K. and School of Computer Science, University of Sheffield, S1 4DP Sheffield, U.K. (e-mail: v.osmani@qmul.ac.uk).}
\thanks{Chen Chen is with School of Computer Science, University of Sheffield, S1 4DP Sheffield, U.K. and Department of Computing, Imperial College London, SW7 2AZ London, U.K. (e-mail: chen.chen2@sheffield.ac.uk).}
}

\maketitle

\begin{abstract}

Pathology and anatomy are two essential groups of semantics in medical data.
Pathology describes what the diseases are, while anatomy explains where the diseases occur.
They describe diseases from different perspectives, providing complementary insights into diseases.
Thus, properly understanding these semantics and their relationships can enhance medical vision-language models (VLMs).
However, pathology and anatomy semantics are usually entangled in medical data, hindering VLMs from explicitly modeling these semantics and their relationships.
To address this challenge, we propose \ourmodel, a novel \textbf{Me}dical \textbf{D}ual-\textbf{S}tream \textbf{L}anguage-\textbf{I}mage \textbf{P}re-training pipeline, to disentangle pathology and anatomy semantics and model the relationships between them. 
We introduce a dual-stream mechanism in \ourmodel\ to explicitly disentangle medical semantics into \emph{pathology-relevant} and \emph{anatomy-relevant} streams and align visual and textual information within each stream.
Furthermore, we propose an interaction modeling module with \emph{prototypical contrastive learning loss} and \emph{intra-image contrastive learning loss} to regularize the relationships between pathology and anatomy semantics.
We apply \ourmodel\ to chest X-ray analysis and conduct comprehensive evaluations with four benchmark datasets: NIH CXR14, RSNA Pneumonia, SIIM-ACR Pneumothorax, and COVIDx CXR-4.
The results demonstrate \ourmodel's superior generalizability and transferability across different scenarios.
The code is available at \url{https://github.com/Shef-AIRE/MeDSLIP}, and the pre-trained model is released at \url{https://huggingface.co/pykale/MeDSLIP}.

\end{abstract}


\section{Introduction}
\label{sec:introduction}

\IEEEPARstart{P}{athology} and anatomy are two essential groups of semantics in medical images and associated reports.
Pathology focuses on the nature and characteristics of diseases, explaining what the abnormalities are.
Anatomy, on the other hand, provides the structural and locational context, describing where these abnormalities occur~\cite{ap-3}.
For instance, in the sentence, ``Opacity is observed on the bilateral lungs, and deformity of posterior ribs is noted,'' ``opacity'' and ``deformity'' are pathology semantics, while ``lungs'' and ``ribs'' are anatomy semantics~\cite{mimic-cxr}.

Pathology and anatomy semantics describe diseases from different perspectives, offering complementary insights into understanding diseases~\cite{ap-1,ap-2}.
Therefore, clearly modeling these semantics and their relationships can significantly enhance disease understanding via medical vision-language models (VLMs)~\cite{review6,review8}, thereby improving performance on key medical tasks such as medical image analysis~\cite{review1,review4}.

However, pathology and anatomy semantics are deeply entangled in medical contexts: 
Pathology semantics are often contextualized within specific anatomical regions.
This entanglement hinders medical VLMs from explicitly understanding pathology and anatomy semantics and their intrinsic relationships, which further leads to the underutilization of the information in data.
Hence, it will be beneficial to disentangle pathology and anatomy semantics from data and model their relationship with explicit guidance.

Most existing medical VLMs for extracting semantics from medical data are text-centric.
They often prioritize textual hierarchy~\cite{biovil,chexzero,convirt,cxr-clip,m-flag,REFERS,med-unic,medclip,gloria,imitate,mgca} and semantics~\cite{rc-tpl,k-diag,kad,medklip} in medical reports with limited exploitation of the intricate visual semantics presented in medical images. 
This imbalance leads to underutilization of pathology and anatomy semantics in medical images, missing the opportunity to use their complementary insights for more effective image analysis.

Two VLM pre-training methods to align visual and textual information are \emph{hierarchical alignment} and \emph{semantic alignment}.
Both are text-centric.
(1) Hierarchical alignment stratifies textual information into different levels and aligns visual features accordingly~\cite{biovil,chexzero,convirt,cxr-clip,m-flag,REFERS,med-unic,medclip,gloria,imitate,mgca}. 
This text stratification is usually based on textual hierarchies of the medical reports, such as syntax~\cite{gloria} or discourse~\cite{imitate} hierarchies.
(2) Semantic alignment focuses on semantic concepts in the data but primarily on textual semantics~\cite{rc-tpl,k-diag,kad,medklip,mavl}. 
It usually extracts key semantics from raw medical reports and enhances those textual semantics with prior knowledge from humans (e.g., domain knowledge~\cite{medklip,rc-tpl,mavl}, knowledge graph~\cite{kad,k-diag}, etc).

Emphasizing texts, most current medical VLMs learn visual semantics automatically in pre-training without explicit guidance, leaving visual semantics not fully disentangled and their relationship indirectly modeled.
This leads to underutilization of visual information in pre-training, motivating the need for a more balanced approach that explicitly incorporates visual semantics to improve model performance.

To explicitly disentangle the pathology and anatomy semantics and properly model their relationships, we propose a semantic vision-language alignment pipeline: \ourmodel, \textbf{Me}dical \textbf{D}ual-\textbf{S}tream \textbf{L}anguage-\textbf{I}mage \textbf{P}re-training, and apply it to chest X-ray analysis. 
\ourmodel\ proposes a) a dual-stream mechanism with a disentanglement module to disentangle intertwined pathology and anatomy semantics in images, and b) an interaction modeling module with two contrastive losses to model the relationships between pathology and anatomy semantics. Our contributions are three-fold:

\emph{Firstly}, our dual-stream mechanism separately encodes pathology and anatomy semantics in both medical images and associated reports.
In text processing, \ourmodel\ extracts pathology and anatomy semantics and prompts them with prior knowledge from humans~\cite{medklip}. 
In image processing, we disentangle pathology and anatomy semantics from raw images using a disentanglement module. 
The disentangled visual and textual semantics are then aligned within the pathology-related and anatomy-related streams.
By disentangling pathology and anatomy semantics and aligning them in separate streams, \ourmodel\ provides a clear understanding of the pathology and anatomy semantics.

\emph{Secondly}, our interaction modeling module exploits a Prototypical Contrastive Loss (ProtoCL) and an Intra-image Contrastive Loss (ICL) to model the relationships between pathology and anatomy semantics.
ProtoCL models semantic interactions by aligning the cross-modal, cross-stream information (e.g., pathology in images and anatomy in texts, anatomy in texts and pathology in images).
ICL models the pathology-anatomy interactions in images by measuring the co-existence of pathology and anatomy semantics.
For example, for a sentence like ``Opacity is observed on the bilateral lungs.'' in the report and its corresponding image, ProtoCL aligns ``opacity'' in the image and ``lung'' in the text, and vice versa, while ICL regularizes the co-existence of ``opacity'' and ``lung'' in the image.
By modeling these interactions, \ourmodel\ captures the rich semantic interactions between pathology and anatomy in both images and texts, leading to a better understanding of the relationships between pathology and anatomy semantics.

\emph{Finally}, to validate \ourmodel's effectiveness, we conduct a comprehensive evaluation of classification, grounding, and segmentation tasks under both zero-shot and fine-tuning settings with NIH CXR14~\cite{cxr14}, RSNA Pneumonia~\cite{rsna}, SIIM-ACR Pneumothorax~\cite{siim-acr}, and COVIDx CXRv4~\cite{covidx} datasets.
The results demonstrate \ourmodel's superior generalizability and transferability.
We also conduct an ablation study and qualitative experiments to demonstrate the effectiveness of \ourmodel\ and the contributions of its key modules.

\begin{figure*}[t]
    \centering
    \includegraphics[width=\textwidth, trim=0.5cm 0.2cm 0.2cm 0.5cm]{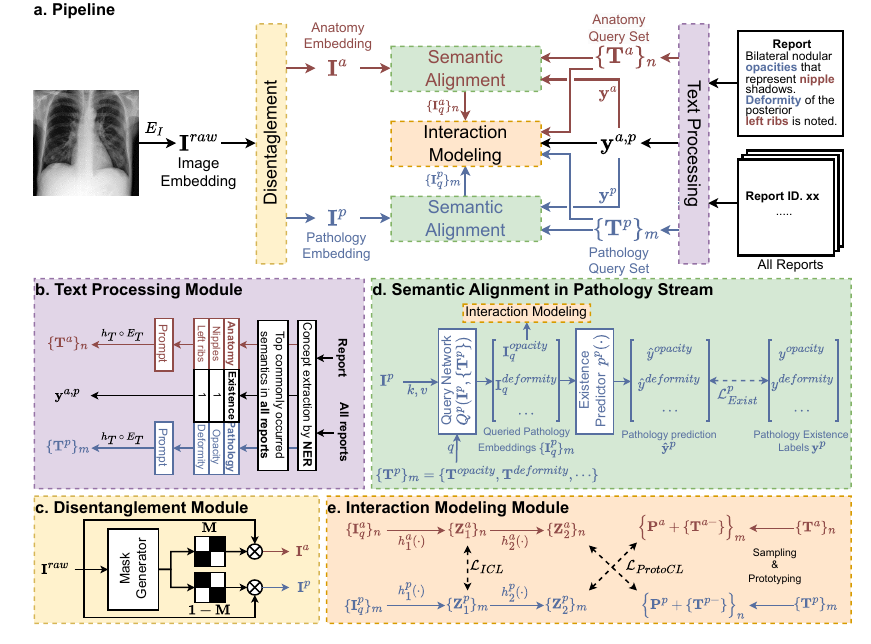}
    \caption{
    Pipeline of \textbf{Me}dical \textbf{D}ual-\textbf{S}tream \textbf{L}anguage-\textbf{I}mage \textbf{P}re-training (MeDSLIP). Each module is indicated with a unique color. Symbols with $\mathbf{I}$ and $\mathbf{T}$ denote image and text embeddings, respectively. $Q$ denotes query networks. $h$ denotes linear projection layers. $\mathbf{Z}$ represents outputs after linear projection. $E_I$ and $E_T$ are image and text encoders, respectively. $\mathbf{y}$ represents existence labels. The denotations with superscripts $p$ and $a$ are pathology-related and anatomy-related.
    \textbf{a. Pipeline:} Reports are processed to extract pathology and anatomy terms, generate text query embedding sets $\easet$ and $\epset$, and an existence label matrix, $\Lap$. 
    $m$ and $n$ represent that we select top commonly seen $m$ pathology semantics and $n$ anatomy semantics in all medical reports.
    Images are encoded, disentangled, and aligned within corresponding streams. 
    The interaction modeling module regularizes the interactions between pathology and anatomy semantics.
    \textbf{b. Text Processing:} (pathology, anatomy, existence) triplets are extracted from raw reports. Most commonly occurring triplets among all reports are used as query sets, which are prompted and encoded to obtain query embeddings (see Sec.~\ref{sec:text-pre-processing}).
    \textbf{c. Disentanglement Module:} It masks raw image embeddings, disentangling pathology and anatomy embedding (Sec.~\ref{sec:disentangle-module}).
    \textbf{d. Semantic Alignment:} A query network $\qp$ aligns the text query set $\epset$ with the image pathology embedding $\rp$ and outputs a queried pathology embedding set $\rpset$. An existence predictor $p^p$ then checks whether each text semantic exists in the images. 
    A similar alignment process is applied to anatomy semantics (see Sec.~\ref{sec:semantical-align}).
    \textbf{e. Interaction Modeling:} $\Licl$ aligns unimodal, cross-stream information, while $\Lproto$ aligns cross-modal, cross-stream information (see Sec.~\ref{section:interaction-modeling}).
    }
    \label{fig:main}
\end{figure*}

\section{Methodology}

Figure~\ref{fig:main} shows \ourmodel's pipeline.
We first disentangle the pathology and anatomy semantics from images and texts and encode them in two distinct streams.
Then, the disentangled visual and textual semantics are aligned within each stream.
An interaction modeling module is proposed to model the relationship between pathology and anatomy semantics.

\subsection{Text Processing}\label{sec:text-pre-processing}

Figure~\ref{fig:main}b shows the text processing in \ourmodel.
We process all medical reports before pre-training in three steps~\cite{medklip}:
(1) Triplet extraction: We extract (pathology, anatomy, existence) triplets from each report~\cite{radgraph}. Then, we select the most commonly occurring triplets in the whole dataset to formulate a pathology query set with $m$ pathology concepts and an anatomy query set with $n$ anatomy concepts.
(2) Prompting: We prompt pathology semantics with domain knowledge derived from professional medical knowledge bases and reliable Internet resources, which provide definitions and explanations of pathology semantics in plain language~\cite {medklip}.
We prompt anatomy semantics by a fixed prompting template: ``It is located at [ANATOMY]''.
(3) Text encoding: We then encode prompted pathology and anatomy query sets by the text encoder, which consists of a frozen pre-trained medical language model~\cite{bioclinicalbert} $E_T$ alongside a learnable linear projection layer $h_T$. 
The generated text embeddings are used as queries in semantic alignment and positive/negative samples in interaction modeling, as depicted in Fig.~\ref{fig:main}d and \ref{fig:main}e.

In this context, we obtain three outputs: a pathology query set $\epset$, an anatomy query set $\easet$, and a set of existence label matrices $\{\Lap\}$.
Two query sets consist of the text embeddings of the top $m$ commonly seen pathology semantics and top $n$ commonly seen anatomy semantics among triplets from all reports in the dataset.
The query sets are universal for all reports.
The existence matrix $\Lap$ is an $n\times m$ matrix, whose element $y^{a_i,p_j}$ indicates whether a pathology observation $\epj$ exists at a specific anatomy location $\eai$.
Columns and rows of $\Lap$ correspond to pathology and anatomy semantics in query sets.
For example, in the sentence ``Deformity of the posterior left ribs is noted'', the pathological observation ``deformity'' exists on the anatomical location ``left ribs''. 
Thus, the existence label $y^{{left\ ribs, deformity}}$ is positive.
Otherwise, the existence label $y^{a,p}$ will be negative if it is not mentioned or is mentioned as not existing.
The meanings of each element in $\Lap$ are universal for all data, but the values are unique for each image.

\subsection{Image Encoding}\label{sec:img-enc}

The image encoding in \ourmodel's pre-training comprises three modules: disentanglement, semantic vision-language alignment, and interaction modeling.
A trainable visual encoder $E_I$ is employed to encode the image into the latent space, where all three modules operate.

\subsubsection{Disentanglement Module}\label{sec:disentangle-module}

We design a disentanglement module to disentangle the intertwined pathology and anatomy semantics in medical images.
As shown in Fig.~\ref{fig:main}c, we use a mask generator, which takes the raw image embedding $\rraw$ as input and outputs a mask $\mathbf{M}$.
$\mathbf{M}$ has the same shape as the input embedding.
The elements in $\mathbf{M}$ range from 0 to 1.
Then, we disentangle pathology and anatomy embeddings $\rp$ and $\ra$ from raw image embedding $\rraw$ through element-wise multiplication with $\mathbf{M}$ and $\mathbf{1-M}$, where $\mathbf{1}$ is an all-ones matrix with the same size as $\mathbf{M}$:
\begin{equation}
    \rp=\mathbf{M} \odot \rraw,\ra=(\mathbf{1}-\mathbf{M})\odot \rraw.
\end{equation}
By disentangling pathology and anatomy semantics into distinct streams, \ourmodel\ decouples the intertwined information about characteristics and locations of the diseases, providing a clearer understanding of the different aspects of diseases.

\subsubsection{Semantic Vision-Language Alignment}\label{sec:semantical-align}

After disentanglement, the pathology and anatomy embeddings $\rp$ and $\ra$ are aligned with corresponding text query embeddings $\ep$ and $\ea$ via query networks $\qp$ and $\qa$, respectively.
Since the semantic alignments in the two streams are similar, we use the pathology stream as an example to illustrate.
As depicted in Fig.~\ref{fig:main}d, the query network $\qp$ takes two inputs: an image embedding $\rp$ and the query set $\epset$.
For each query embedding ${\ep}$ in $\epset$, $\qp$ extracts a queried image embedding $\rpq$ from $\rp$: $\rpq=\qp(\rp,\ep)$.
Here, $\rpq$ represents the corresponding features in $\rp$ related to the specific query $\ep$.
Then, $\rpq$ is passed through a binary existence predictor $p^p(\cdot)$, and we obtain a prediction $\hat{y}^p$ by $\hat{y}^p=p^p(\rpq)$.
The prediction $\hat{y}^p$ tells whether a specific pathology semantic exists in the image.
Repeating the above process for all queries in $\epset$, we obtain a set of pathology predictions $\yhatmatrix^p$.
Finally, we extract the pathology existence labels $\Lp$ from $\Lap$ to calculate the pathology existence loss $\Lexist^p$, a binary cross-entropy loss between prediction $\yhatmatrix^p$ and labels $\Lp$:
\begin{equation}
    \Lexist^p=\mathcal{L}_{BCE}(p^p(\qp(\rp,\ep),\Lp)).
\end{equation}

\subsubsection{Interaction Modeling}\label{section:interaction-modeling}

To properly model the interactions between pathology and anatomy semantics and produce a unified output that retains all relevant information for downstream tasks, we propose an interaction modeling module, as shown in Fig.~\ref{fig:main}e.
This module regularizes the interactions between visual information from one stream and both visual and textual information from the other stream by two specialized losses: prototypical contrastive loss (ProtoCL) and intra-image contrastive loss (ICL). 
ProtoCL emphasizes the interactions between image embeddings in one stream and text embeddings in the other, while ICL focuses on image embeddings between the two streams.

\begin{figure}[t]
    \centering
    \includegraphics[width=0.8\linewidth, trim=1cm 0cm 0cm 0cm]{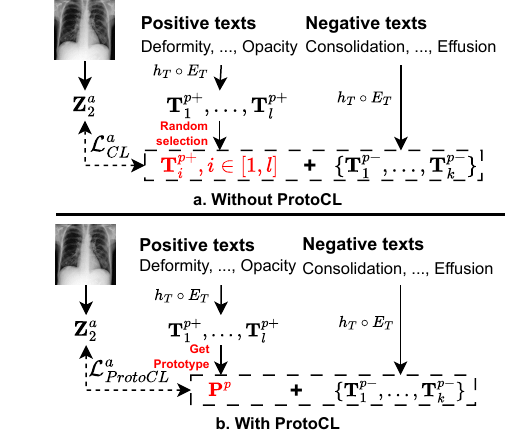}
    \caption{Comparison between contrastive learning with or without prototypes, using ProtoCL between anatomy image embeddings and pathology text embeddings as an example. \textbf{a.} Conventional contrastive learning without prototypes. \textbf{b.} ProtoCL uses the prototype of all positive samples as the new positive example in contrastive learning.}
    \label{fig:protocl}
\end{figure}

\paragraph{Prototypical Contrastive Loss (ProtoCL)} ProtoCL regularizes the interactions between cross-modal, cross-stream information by aligning image embeddings in one stream with text embeddings from the other. 
As is shown in Fig.~\ref{fig:protocl}, we use ProtoCL between anatomy image embeddings and pathology text embeddings as an example to illustrate how ProtoCL works and highlight the difference between ProtoCL and conventional contrastive learning.

For medical images and radiology reports, we usually regard existing pathology or anatomy semantics as positive and other unmentioned or non-existing semantics as negative.
In this context, there are often multiple positive examples due to the possible coexisting diseases.
ProtoCL employs the prototype of all positive examples as the new positive example. 
The prototype is the center of all positive samples in the textual embedding space, which is calculated by 
\begin{equation}
    \mathbf{P}=\frac{1}{l}\sum_{i=0}^{l}{\mathbf{T}_i^{+}}.
    \label{eq:proto}
\end{equation}
We use a Noise Contrastive Estimation (NCE) loss~\cite{nce} between prototypes and sampled negatives in ProtoCL, which is calculated as follows:
\begin{equation}
    \Lproto^a=-\mathbb{E}\Biggl[
        log\frac{exp(\zab \cdot \mathbf{P}^p)}{\Sigma_{i=1}^{k}exp(\zab \cdot \mathbf{T}_i^{p-})}
    \Biggr].
    \label{eq:protocl}
\end{equation}

Although conventional contrastive learning can also model the interactions, ProtoCL's key contribution is using prototypes as positive examples.
It is motivated by the underutilization of positive information in conventional contrastive learning.
When faced with multiple positive examples, conventional contrastive learning tends to randomly select one of the positive examples and leave others unused to keep a low positive-negative ratio~\cite{medklip} for NCE loss, leading to the underutilization of information in the unselected positives.
Instead, we design ProtoCL to aggregate all positive information in prototypes without increasing the positive-negative ratio.
Therefore, in addition to modeling cross-modal, cross-stream interactions, ProtoCL improves the data efficiency of positive examples by using all positive instances without increasing the positive-to-negative ratio.

\begin{figure}[t]
    \centering
    \includegraphics[width=\linewidth, trim=0cm 0cm 0.5cm 0cm]{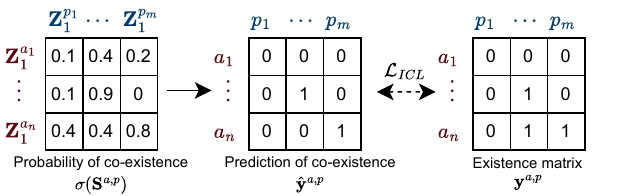}
    \caption{Mechanism of intra-image contrastive loss (ICL). $a_1, \dots, a_n$ and $p_1, \dots, p_m$ indicating the semantics in query sets. Elements in $\yhatmatrix^{a,p}$ are the predictions, and elements in $\Lap$ are ground truths when computing $\Licl$.}
    \label{fig:icl}
\end{figure}

\paragraph{Intra-image Contrastive Loss (ICL)} ICL regularizes the visual semantics across two streams by measuring the co-existence of (pathology, anatomy) pairs.
A (pathology, anatomy) pair is considered positive if a specific pathology observation is present at a corresponding anatomy location ($\lap=1$).
Otherwise, the pair is treated as negative ($\lap=0$).
As depicted in Fig.~\ref{fig:icl}, we first compute a cosine similarity matrix $\Sap$ between the queried anatomy image embeddings $\raset$ and pathology image embeddings $\rpset$. The element $s^{a_i,p_j}$ of $\Sap$ is given by 
\begin{equation}
    s^{a_i,p_j}=\langle\mathbf{Z}^{a_i}_1, \mathbf{Z}^{p_j}_1\rangle, i\in [1,n], j\in [1,m],
\end{equation}
where $\zaa$ and $\zpa$ represent linearly projected image embeddings corresponding to specific semantics in the texts.
The similarity matrix $\Sap$ is then passed through a sigmoid activation layer $\sigma(\cdot)$ to produce a probability matrix $\sigma(\Sap)$, which represents the likelihood of co-existence for each (pathology, anatomy) pair. 
Then, we obtain predictions of co-existence $\yhatmatrix^{a,p}$ from the probability matrix $\sigma(\Sap)$.
We use the existence matrix $\Lap=\{l^{a_i,p_j}\}, i\in [1,n], j\in[1,m]$ from the extracted (pathology, anatomy, existence) triplets as ground truths to compute the ICL loss $\Licl$.
Each element $\lap$ in $\Lap$ serves as the existence label for the corresponding (pathology, anatomy) pair.
Finally, $\Licl$ is obtained by a binary cross-entropy (BCE) loss between $\yhatmatrix^{a,p}$ and $\Lap$:
\begin{equation}
    \Licl=\mathcal{L}_{BCE}(\yhatmatrix^{a,p}, \Lap).
    \label{eq:icl}
\end{equation}
The ICL loss encourages the alignment of pathology and anatomy embeddings for positive (pathology, anatomy) pairs while minimizing alignment for negative pairs. 
This ensures that the model captures fine-grained relationships in images between anatomical structures and pathological conditions, further enhancing interaction modeling from both streams.

In summary, the total loss function in pre-training is
\begin{equation}
\mathcal{L}=\Lexist+\alpha\Lproto+\beta\mathcal{L}_{ICL},
\label{eq:total-loss}
\end{equation}
where $\alpha$ and $\beta$ are temperature coefficients that balance the scale of the different loss components. 
$\Lexist$ and $\Lproto$ represent the summed losses from the pathology and anatomy streams for existence prediction and prototypical contrastive learning, respectively.

\subsection{Inference} 
The existence predictor equips the model with zero-shot classification capability. 
For instance, when encountering an unseen disease, we prompt the disease in a similar way to how seen pathology semantics are handled during pre-training. 
The prompted semantics are then encoded by the same text encoder and used as a query in the query network. 
After obtaining the queried image embedding corresponding to the disease, the existence predictor determines whether the disease is present in the image.
Since most downstream tasks are focused on pathology information, we use embeddings in the pathology stream as the model's outputs for downstream tasks ($\rpset$ and $\yhatmatrix^p$ for zero-shot tasks, $\rraw$ for fine-tuning tasks).

\section{Experiment Settings}
\subsection{Datasets}

\begin{table}[t]
\centering
\caption{Metadata of datasets used in experiments.}
\label{tab:metadata}
\resizebox{\linewidth}{!}{%
\begin{tabular}{llrl}
\toprule
\toprule
Stage & Dataset & \# Images
& Disease \\ \midrule
Pre-training & MIMIC-CXR~\cite{mimic-cxr,mimic-cxr-base,mimic-cxr-jpg} & 377,110 & - \\ \midrule
\multirow{6}{*}{Evaluation} & NIH CXR14~\cite{cxr14} & 112,000 & 14 diseases \\ \cmidrule{2-4} 
 & RSNA Pneumonia~\cite{rsna} & 30,000 & Pneumonia \\ \cmidrule{2-4} 
 & \begin{tabular}[c]{@{}l@{}}SIIM-ACR Pneumothorax~\cite{siim-acr}\end{tabular} & 12,047 & Pneumothorax \\ \cmidrule{2-4} 
 & COVIDx CXR-4~\cite{covidx} & 84,818 & COVID-19 \\
 \bottomrule
 \bottomrule
\end{tabular}%
}
\end{table}

We use MIMIC-CXR~\cite{mimic-cxr,mimic-cxr-base,mimic-cxr-jpg} for pre-training. NIH CXR14~\cite{cxr14}, RSNA Pneumonia~\cite{rsna}, SIIM-ACR Pneumothorax~\cite{siim-acr}, and COVIDx CXR-4~\cite{covidx} are involved in evaluation.
The metadata of these datasets is shown in TABLE~\ref{tab:metadata}.

\subsection{Implementation}
\begin{table}[]
\centering
\caption{Hyperparameters in \ourmodel\ pre-training and fine-tuning. The different values in fine-tuning indicate hyperparameters in classification/segmentation fine-tuning tasks.}\label{tab:hyper}
\begin{tabular}{lrr}
\toprule
\toprule
Name & Pre-training & Fine-tuning \\ \midrule
Hidden size & 256 & 256 \\
Image size & 224$\times$224 & 224$\times$224 \\
Batch size & 64 & 64\\
$n$ & 50 & 50 \\
$m$ & 75 & 75\\
($\alpha$, $\beta$) & (1, 1) & - \\
Optimizer & AdamW & AdamW\\
Epoch & 100 & 100/1000\\
Peak learning rate & 1e-4 & 1e-4/1e-5\\
Min learning rate & 1e-5 & 1e-5\\
Warmup epoch & 5 & 20/50\\
Warmup learning rate & 1e-5 & 1e-5\\
Weight decay & 0.02 & 5e-4 \\
Scheduler & cosine annealing & cosine annealing \\ 
Time &  $\sim$ 2 days & minutes to hours \\
\midrule
GPU &  \multicolumn{2}{r}{NVIDIA GeForce RTX4090 24GB} \\
\bottomrule
\bottomrule
\end{tabular}%
\end{table}
To demonstrate the feasibility of our proposed pipeline, we choose simple structures for each module.
\ourmodel\ uses ResNet-50~\cite{resnet} as the image encoder and Bio-ClinicalBERT~\cite{bioclinicalbert} with a learnable single-layer perceptron (SLP) as the text encoder.
For the disentanglement module, an SLP layer is used as the mask generator.
We list hyperparameters of \ourmodel\ in pre-training and downstream tasks in TABLE~\ref{tab:hyper}.

\begin{table*}[t]
\centering
\caption{Zero-shot classification evaluation with SOTA CNN-based models. For CXR14, metrics refer to the macro average on the 14 diseases.}
\label{table:zs-cls}
\begin{threeparttable}
\begin{tabular}{lcccccccccccc}
\toprule
\toprule
\multicolumn{1}{c}{\multirow{2}{*}{Models}} & \multicolumn{3}{c}{NIH CXR14~\cite{cxr14}}     & \multicolumn{3}{c}{RSNA Pneumonia~\cite{rsna}} & \multicolumn{3}{c}{SIIM-ACR Pneumothorax~\cite{siim-acr}}    & \multicolumn{3}{c}{COVIDx CXR-4~\cite{covidx}}            \\ \cmidrule{2-13} 
   & AUROC$\uparrow$  & F1$\uparrow$   & ACC$\uparrow$      & AUROC$\uparrow$            & F1$\uparrow$             & ACC$\uparrow$    & AUROC$\uparrow$  & F1$\uparrow$   & ACC$\uparrow$  & AUROC$\uparrow$  & F1$\uparrow$   & ACC$\uparrow$      \\
\midrule
GLoRIA\textsuperscript{\ding{58}}~\cite{gloria}                 & 66.10          & 17.32          & 77.00       & 71.45          & 49.01          & 71.29  & 53.42  & 38.23  & 40.47 & - & - & - \\
MedCLIP~\cite{medclip} & 62.91 & 14.29 & 54.28 & 81.66 & 56.77 & 73.77 & 67.94 & 45.51 & 61.03 & 68.84 & \underline{69.60} & 64.57 \\
ConVIRT~\cite{convirt}                & 65.78          & 15.62          & 64.83       & 81.54          & 56.83          & 75.83  & 68.73  & 45.66 & 55.53 & 66.06 & 66.71 & 50.04        \\
BioViL~\cite{biovil}                 & 73.92          & 23.56          & 79.62       & 82.54          & 56.62          & 76.73  & 73.53  & 50.63  & 65.15 & 69.82 & 67.20 & 54.41  \\
CheXzero~\cite{chexzero}               & 75.65          & 22.67          & \underline{86.22}       & \underline{86.11}          & 55.43    & 66.32  & 77.28  & 47.94  & 56.24 & 71.56 & 60.84 & 69.78    \\
UniChest\textsuperscript{\ding{61}}~\cite{unichest} & - & - & - & 79.04 & 53.55 & 70.32 & 89.20 & \underline{56.52} & 83.71 & 68.64 & 67.47 & 55.01 \\
CXR-CLIP~\cite{cxr-clip}               & 75.49              & 18.17              & 66.71           & 82.91          & 57.57          & 72.93    & 84.02              & 50.10              & 57.36 & 71.88 & 62.57 & 68.38      \\
CXR-CLIP\textsuperscript{\ding{61}}~\cite{cxr-clip}               & -              & -              & -           & 83.41          & 53.80          & 64.39   & 85.86 & 42.28 & 40.15 & 54.17 & 52.38 & 57.05 \\
MedKLIP~\cite{medklip} & \underline{78.00}    & \underline{25.71}    & 85.22 & 85.27    & \underline{61.13}          & \underline{79.91}   & \underline{89.60}  & {56.39}  & \underline{84.84} & \underline{82.77} & 66.82 & \underline{73.09} \\

\midrule
\ourmodel\ (Ours)               & \textbf{80.34} & \textbf{29.96} & \textbf{88.93}    & \textbf{86.49} & \textbf{63.81} & \textbf{80.98} & \textbf{89.70}  & \textbf{58.22}  & \textbf{85.05} & \textbf{83.41} & \textbf{77.15} & \textbf{77.21}\\
\bottomrule
\bottomrule
\end{tabular}

\begin{tablenotes}
\item \ding{58} Because GLoRIA is trained on in-house data, we quote its results in \cite{medklip}. Because we use a different version of COVIDx~\cite{covidx} than~\cite{medklip}, we don't report GLoRIA's results on COVIDx.
\ding{61} They include CXR14 as a part of pre-training data. Thus, we don't perform the zero-shot classification task on CXR14.
\end{tablenotes}

\end{threeparttable}
\end{table*}

\subsection{Baselines}
We compare \ourmodel\ to eight leading CNN-based models in the field: ConVIRT~\cite{convirt}, MedCLIP~\cite{medclip}, GLoRIA~\cite{gloria}, BioViL~\cite{biovil}, CheXzero~\cite{chexzero}, UniChest~\cite{unichest}, MedKLIP~\cite{medklip}, and CXR-CLIP~\cite{cxr-clip}. 
Since GLoRIA~\cite{gloria}is trained on an in-house dataset, we quote the results in \cite{medklip}. 
For BioViL~\cite{biovil}, CheXzero~\cite{chexzero}, UniChest~\cite{unichest}, and CXR-CLIP~\cite{cxr-clip}, we use the officially released pre-trained weights.
MedCLIP~\cite{medclip} doesn't release its code, so we reproduce it according to its paper and pre-train it under our hyperparameters.
Besides, since MedKLIP~\cite{medklip} doesn't release pre-trained weights, we pre-train it with their official code.

\subsection{Metrics}
We report the area under the ROC curve (AUROC), F1 score, and accuracy (ACC) for classification tasks, intersection over union (IoU), Dice coefficient, and pointing game score (PG)~\cite{medklip} for grounding tasks, and Dice coefficient for segmentation tasks.
Besides, we use $t$-SNE distribution and UMAP to visualize the feature space.

\section{Experiment Results}

To assess \ourmodel's capabilities as a vision-language pre-training framework, we evaluate its generalizability and transferability on classification, grounding, and segmentation tasks under both zero-shot and fine-tuning settings. 
Additionally, we conduct a comprehensive ablation study and validate the effectiveness of each proposed module.
Upward arrows ($\uparrow$) of metrics indicate that higher values are better, and downward arrows ($\downarrow$) indicate that lower values are preferred. 
The best results are highlighted in \textbf{bold}, while the second-best results are \underline{underlined}. We format all results in tables in percentages.

\subsection{Generalizability Evaluation}


\begin{figure*}[t]
    \centering
    \includegraphics[width=1\textwidth, trim=0cm 0cm 0cm 0cm]{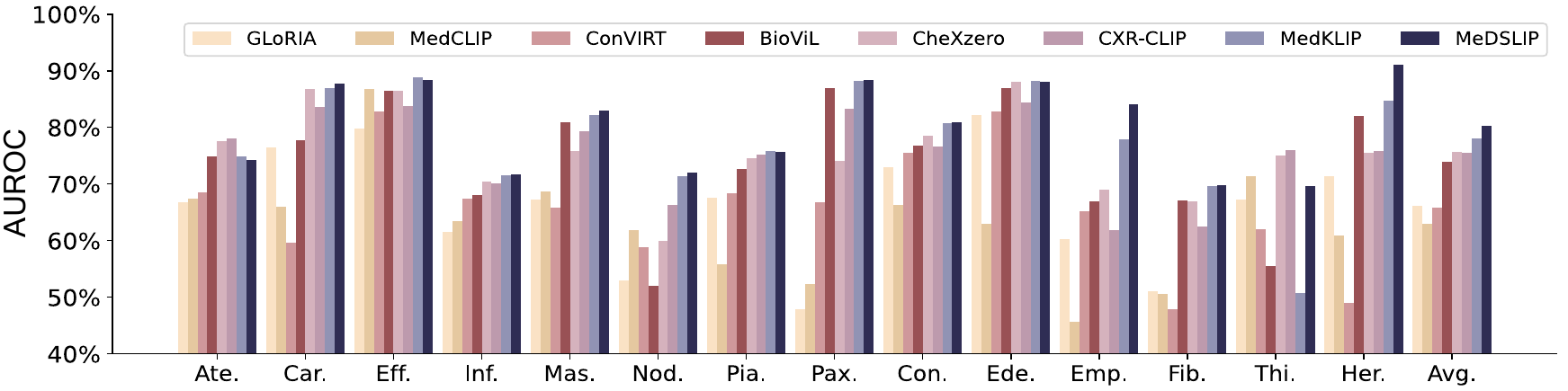}
    \caption{Disease-wise AUROCs of zero-shot classification on NIH CXR14 dataset~\cite{cxr14} show \ourmodel\ outperforms other baselines on most of the diseases. AUROCs are calculated between the positive patients of each disease and other health controls across all data.}
    \label{fig:classwise-auc}
\end{figure*}

\begin{figure*}[t]
    \centering
    \includegraphics[width=1\textwidth, trim=0cm 0cm 0cm 1cm]{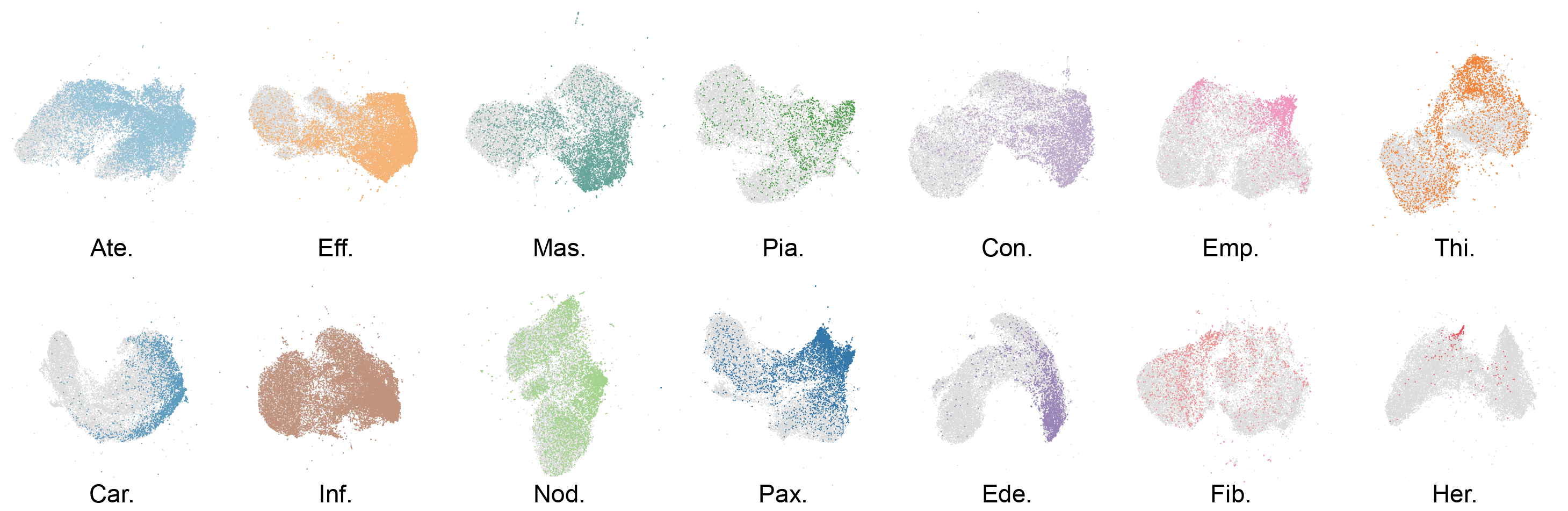}
    \caption{Disease-wise UMAPs of $\{\rpq\}_{14}$ of NIH CXR14 dataset~\cite{cxr14}. Gray points in each UMAP represent $\rpq$ of healthy controls, while colored points denote $\rpq$ of patients with the corresponding disease. The diseases with higher AUROC scores in TABLE~\ref{table:zs-cls} tend to be more distinct and well-clustered.}
    \label{fig:classwise-umap}
\end{figure*}

\begin{figure}[t]
    \centering
    \includegraphics[width=\linewidth, trim=0cm 0cm 0cm 0cm]{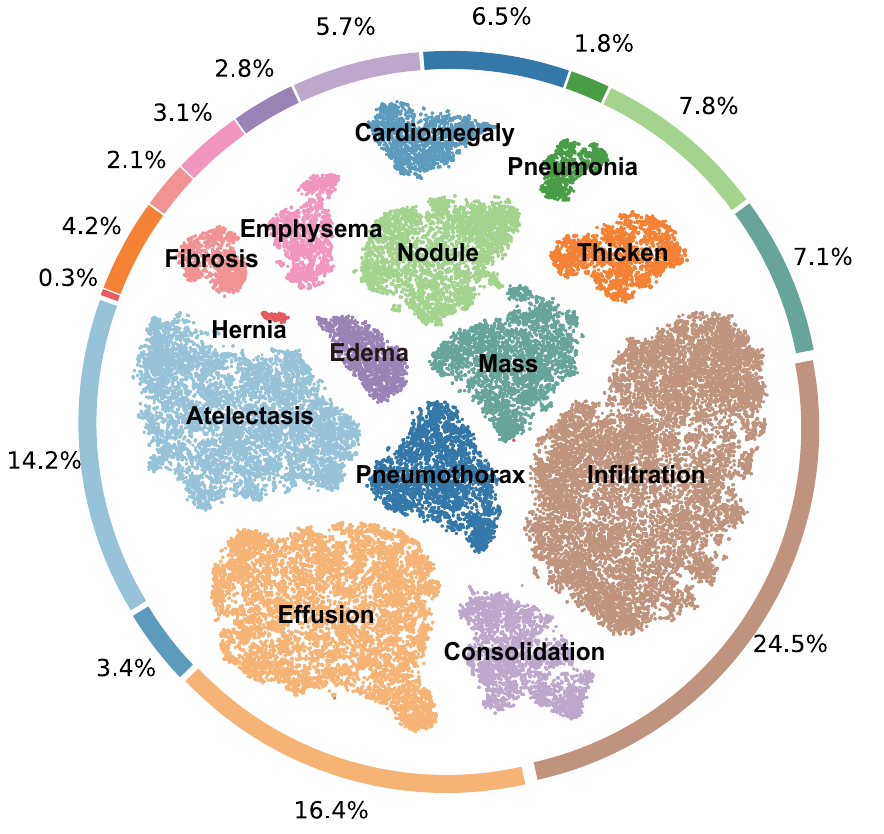}
    \caption{MeDSLIP clearly distinguishes different pathology semantics.
    The center of the figure is the $t$-SNE distribution of queried pathology embeddings $\rpset$ to 14 diseases. 
    The outside donut chart shows the class distribution in the NIH CXR14 dataset~\cite{cxr14}.}
    \label{fig:cxr-14}
\end{figure}

\subsubsection{Zero-shot Classification}
We first demonstrate the strong generalizability of \ourmodel\ with zero-shot classification tasks.
Our zero-shot classification evaluation is conducted in two settings: unseen datasets with seen diseases, and unseen diseases.
Unseen datasets with seen diseases indicate that the evaluation datasets are not seen in the model pre-training, but the diseases are seen.
Unseen disease indicates that the types of diseases in the evaluation dataset are novel for the model.
\paragraph{Zero-shot Classification on Seen Diseases}

We use NIH CXR14~\cite{cxr14}, RSNA Pneumonia~\cite{rsna}, and SIIM-ACR Pneumothorax~\cite{siim-acr} for the zero-shot classification of seen diseases but unseen datasets. 
In this experiment, although the diseases in the evaluation datasets are included in the pre-training dataset (MIMIC-CXR), the data are novel to \ourmodel.
The results are reported in TABLE~\ref{table:zs-cls}. 
\ourmodel\ demonstrates strong efficacy and generalizability in zero-shot classification on unseen datasets, outperforming all other baselines across all datasets and metrics by a significant margin. 

Because NIH CXR14~\cite{cxr14} contains multiple diseases, including pneumonia and pneumothorax, we use it for visualization and in-depth analysis.
NIH CXR14~\cite{cxr14} dataset is a multi-label classification dataset and consists of 14 binary classification problems.

We first present a disease-wise analysis.
Figure~\ref{fig:classwise-auc} shows disease-wise AUROC scores for \ourmodel\ and seven other baselines.
Figure~\ref{fig:classwise-umap} shows disease-wise UMAPs of $\rpq$ between patients of a specific disease (colored points) and healthy controls (gray points).
From the disease-wise AUROC bar graph, \ourmodel\ outperforms other baselines for most conditions. 
Particularly, it significantly improves performance for diseases with traditionally low AUROC scores in baselines, such as emphysema (Emp.) and hernia (Her.). 
This demonstrates \ourmodel's ability to handle a variety of cardiovascular diseases in chest X-rays.
For diseases such as consolidation (Con.), cardiomegaly (Car.), and effusion (Eff.), the UMAPs show less overlap between diseased and healthy embeddings. 
This corresponds to higher AUROC scores, confirming that better feature separation correlates with better classification performance.

During the evaluation, 14 pathology embeddings $\{\rpq\}_{14}$ are generated by the pathology query network $\qp$ for each image, with each embedding $\rpq$ corresponding to one particular disease.
Therefore, we visualize the feature space of $\{\rpq\}_{14}$ to further explore how well \ourmodel\ is to recognize different diseases.
Figure~\ref{fig:cxr-14} illustrates the label distribution of the dataset and the $t$-SNE distribution of the queried pathology embeddings $\{\rpq\}_{14}$.
The $t$-SNE visualization shows well-clustered embeddings with clear margins between diseases, indicating that \ourmodel\ learns a robust feature space even for unseen data distributions during pre-training.

\paragraph{Zero-shot Classification on Unseen Diseases}

We evaluate \ourmodel's zero-shot classification ability on unseen diseases using the COVIDx CXR-4~\cite{covidx} dataset. 
The pre-training dataset, MIMIC-CXR~\cite{mimic-cxr}, collected data between 2011 and 2016.
It predated the COVID-19 pandemic, which began in 2019. 
Therefore, COVID-19 is a novel disease class for evaluating \ourmodel.

Since COVID-19 is not included in previous prompts, we designed a prompt: 
``COVID-19, caused by the SARS-CoV-2 virus, primarily affects the respiratory system and can be identified on chest X-rays by characteristic bilateral, peripheral ground-glass opacities. 
These radiographic findings are most commonly seen in the lower lobes of the lungs and can progress to multifocal consolidation as the disease advances.''


We present the results of zero-shot classification on the COVIDx CXR-4 dataset~\cite{covidx} in TABLE~\ref{table:zs-cls}. 
In the zero-shot classification task, \ourmodel\ consistently outperforms all state-of-the-art (SOTA) methods across all metrics. 
Notably, \ourmodel\ improves accuracy by at least 4.12\% and increases the F1 score by 7.55\%, demonstrating its robustness and superior generalizability in handling unseen diseases.

\begin{table}[t]
\centering
\caption{Zero-shot grounding tasks on RSNA Pneumonia dataset~\cite{rsna}.}
\label{table:zs-grounding}
\begin{threeparttable}
\begin{tabular}{lccc}
\toprule\toprule
\multicolumn{1}{c}{Models} & Dice$\uparrow$ & IoU$\uparrow$  & PG$\uparrow$   \\ \midrule
GLoRIA~\cite{gloria}                      & 34.68          & 21.82          & 76.07          \\
BioViL~\cite{biovil}                      & 44.34          & 30.76          & 84.12          \\
MedKLIP~\cite{medklip}                     & {\underline{49.63}}    & {\underline{34.32}}    & {\underline{87.02}}    \\ \midrule
\ourmodel\                     & \textbf{50.60} & \textbf{35.47} & \textbf{91.10} \\ \bottomrule\bottomrule
\end{tabular}
\end{threeparttable}
\end{table}

\subsubsection{Zero-shot Grounding}

\ourmodel's grounding ability is evaluated through a zero-shot grounding task on the RSNA Pneumonia dataset~\cite{rsna}. 
We use attention maps from the pathology query network $\qp$ and a predefined threshold to identify the abnormal regions. 
Regions with attention values exceeding the threshold are regarded as abnormal regions, which are then compared against the ground truth to compute the evaluation metrics.

Four models are included in this experiment, with results presented in TABLE~\ref{table:zs-grounding}.
\ourmodel\ outperforms all baselines across the three metrics, demonstrating its strong grounding performance. 
It achieves the highest Dice coefficient of 50.60\%, indicating superior overlap between predicted regions and ground truths.  
The model also records the highest IoU of 35.47\%, demonstrating better performance in grounding tasks.
Especially, \ourmodel\ achieves a Pointing Game Score of 91.10\%, significantly outperforming the other models. 
This result highlights \ourmodel's superior capability to localize objects within the images. 
Together, these findings confirm \ourmodel's overall superiority in zero-shot grounding tasks compared to competing methods.

\begin{figure}[t]
    \centering
    \includegraphics[width=\linewidth, trim=0cm 0cm 0cm 0cm]{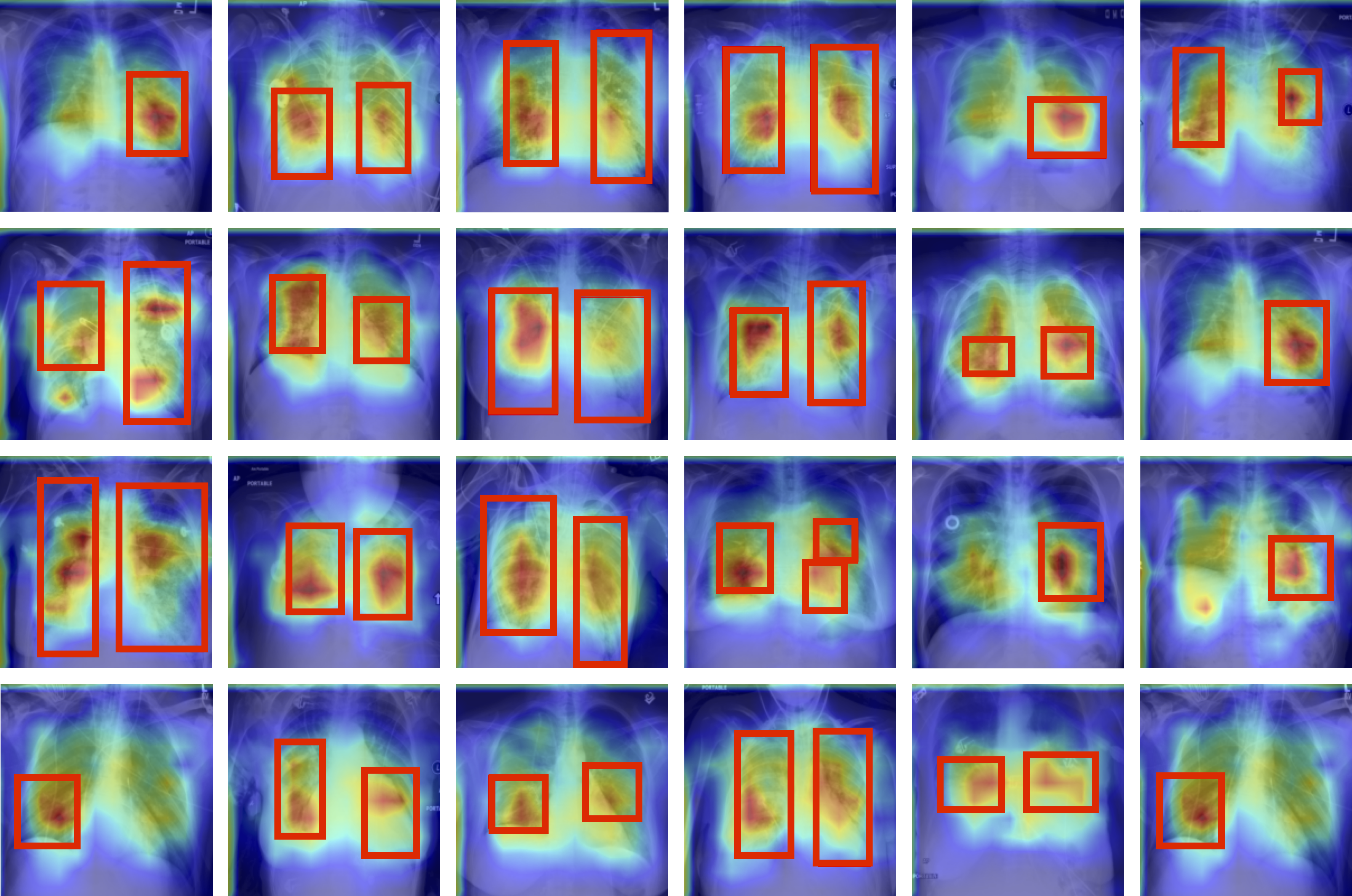}
    \caption{Visualization of attention maps and ground truths of zero-shot grounding task on RSNA Pneumonia dataset~\cite{rsna} shows that \ourmodel\ identifies abnormal regions. The red boxes are ground truths.}
    \label{fig:visual-rsna}
\end{figure}

We visualize attention maps alongside annotated bounding boxes in Fig.~\ref{fig:visual-rsna}. 
In pneumonia grounding tasks, \ourmodel\ first localizes the lungs, assigning them more attention than surrounding areas. 
It then highlights abnormal regions within the lungs with significantly stronger attention responses.

\begin{table}[]
\centering
\caption{Fine-tuning on classification and segmentation tasks.}
\label{table:FT}
\resizebox{\linewidth}{!}{%
\begin{tabular}{clrcc}
\toprule
\toprule
\multirow{2}{*}{Task} & \multicolumn{1}{c}{\multirow{2}{*}{Dataset}} & \multicolumn{3}{c}{Model} \\\cmidrule{3-5}
 & \multicolumn{1}{c}{} & \multicolumn{1}{c}{Data Ratio} & \multicolumn{1}{c}{MedKLIP~\cite{medklip}} & \multicolumn{1}{c}{MeDSLIP} \\\midrule
\multirow{9}{*}{\begin{tabular}[c]{@{}c@{}}Classification\\ (AUROC$\uparrow$)\end{tabular}} & \multirow{3}{*}{CXR14~\cite{cxr14}} & 1\% & 66.20 & \textbf{70.60} \\
 &  & 10\% & 76.66 & \textbf{77.20} \\
 &  & 100\% & 80.30 & \textbf{80.50} \\\cmidrule{2-5}
 & \multirow{3}{*}{SIIM-ACR~\cite{siim-acr}} & 1\% & 86.27 & \textbf{88.06} \\
 &  & 10\% & 89.90 & \textbf{90.96} \\
 &  & 100\% & 93.10 & \textbf{93.83} \\\cmidrule{2-5}
 & \multirow{3}{*}{COVIDx~\cite{covidx}} & 1\% & 65.02 & \textbf{67.41} \\
 &  & 10\% & 74.47 & \textbf{74.64} \\
 &  & 100\% & 77.46 & \textbf{80.38} \\\midrule
\multirow{3}{*}{\begin{tabular}[c]{@{}c@{}}Segmentation\\ (Dice$\uparrow$)\end{tabular}} & \multirow{3}{*}{SIIM-ACR~\cite{siim-acr}} & 1\% & 67.71 & \textbf{70.39} \\
 &  & 10\% & 75.52 & \textbf{77.53} \\
 &  & 100\% & 76.96 & \textbf{77.75} \\
 \bottomrule
 \bottomrule
\end{tabular}%
}
\end{table}

\begin{table*}[t]
\centering
\begin{threeparttable}
\caption{Ablation study under zero-shot setting: BL, PCL, ICL, DIS, represent baseline (MedKLIP), ProtoCL loss, ICL loss, and disentanglement module with dual-stream structure and mask generator.}
\label{table:ablation}
\begin{tabular}{cccccccccccc}
\toprule
\toprule
\multirow{4}{*}{Exp. ID} & \multicolumn{4}{c}{\multirow{2}{*}{Modules}}      & \multicolumn{6}{c}{Classification}          & Grounding\textsuperscript{\ding{169}}                 \\ \cmidrule{6-12} 
                    & \multicolumn{4}{c}{} & \multicolumn{3}{c}{CXR14\textsuperscript{\ding{61}}~\cite{cxr14}}& \multicolumn{3}{c}{RSNA~\cite{rsna}}    & RSNA~\cite{rsna}\\ \cmidrule{2-5}\cmidrule{6-11}\cmidrule{12-12} 
  & BL         & DIS        & PCL        & ICL        & AUROC$\uparrow$  & F1$\uparrow$ & ACC$\uparrow$ & AUROC$\uparrow$ & F1$\uparrow$ & ACC$\uparrow$ & Point Score$\uparrow$ \\ \midrule
1 & \Checkmark &            &            &            & 76.15          & 22.98        & 84.08         & 85.74         & 61.88        & 79.76          & 87.02          \\
2 & \Checkmark & \Checkmark &            &            & 77.30          & 25.28        & 85.00         & 85.62         & 62.13        & 78.60          & 82.45 \\
3 & \Checkmark &            & \Checkmark &            & 77.24          & 23.42          & 86.68          & \underline{85.77}          & \underline{62.57}          & \underline{81.51}          & \underline{87.69}          \\
4 & \Checkmark & \Checkmark & \Checkmark &            & \underline{77.81}          & \textbf{26.59} & \underline{86.90}          & 84.99          & 60.64          & 80.08          & 83.98 \\
5 & \Checkmark & \Checkmark & \Checkmark & \Checkmark & \textbf{78.14} & \underline{26.01}          & \multicolumn{1}{c}{\textbf{88.88}} & \textbf{86.49} & \textbf{63.18} & \textbf{82.97} & \textbf{91.10} \\
\bottomrule
\bottomrule
\end{tabular}

\begin{tablenotes}
\item \ding{169} We use the hyperparameter-agnostic pointing game score~\cite{pointgame} in the grounding task to avoid the effects of hyperparameter selection. 
\ding{61} The study on CXR14~\cite{cxr14} uses the full dataset.
\end{tablenotes}
\end{threeparttable}
\end{table*}


\subsection{Transferability Evaluation}

\subsubsection{Fine-tuning Classification}

We evaluate \ourmodel's transferability on fine-tuning classification tasks using the NIH CXR14~\cite{cxr14}, SIIM-ACR Pneumothorax~\cite{siim-acr}, and COVIDx CXR-4~\cite{covidx} datasets. 
Classification models typically comprise a feature extractor to generate embeddings and a classification head to predict.
We employ the pre-trained image encoder of \ourmodel\ as the feature extractor and a randomly initialized binary classifier as the classification head. 
During fine-tuning, the encoder is frozen while the classifier is trainable.
The model is fine-tuned using 1\%, 10\%, and 100\% of the training set. 
The AUROC scores are reported in TABLE~\ref{table:FT}. 
Across different ratios of training data, \ourmodel\ consistently achieves higher AUROC scores than the baseline, demonstrating its superiority in fine-tuning classification tasks.

Notably, \ourmodel\ shows a significant advantage when trained on the smallest data size, highlighting its data efficiency with limited data. While the performance gap narrows as the training data size increases, \ourmodel\ still maintains a clear lead, underscoring its robustness and adaptability in classification tasks, even in data-rich scenarios.

\subsubsection{Fine-tuning Segmentation}
We further evaluate \ourmodel\ on fine-tuning segmentation tasks using the SIIM-ACR Pneumothorax dataset~\cite{siim-acr}. 
Segmentation models typically comprise an encoder to extract features from raw images and a pixel-dense decoder to generate segmentation maps. 
For this experiment, we use ResUNet~\cite{resunet} as the backbone network. 
The pre-trained image encoder from \ourmodel\ is employed to initialize the encoder of ResUNet, while the decoder is randomly initialized. 
During fine-tuning, the encoder remains frozen, and only the decoder is trainable.
The training data sizes used for segmentation experiments mirror those of the fine-tuning classification tasks, with 1\%, 10\%, and 100\% of the SIIM-ACR Pneumothorax dataset~\cite{siim-acr}. 
The dice scores for these experiments are presented in TABLE~\ref{table:FT}.

The results reveal a consistent trend similar to that observed in the classification tasks. 
Across all training data sizes, \ourmodel\ outperforms the baseline. 
Notably, the performance improvement of \ourmodel\ is most pronounced when less training data is available, demonstrating its high data efficiency with limited data. 
These findings further illustrate the robustness and superiority of \ourmodel\ in fine-tuning segmentation tasks, particularly in data-scarce scenarios.

\subsection{Ablation Study}
To show the contributions of each module in \ourmodel, we perform an ablation study, with results presented in TABLE~\ref{table:ablation}.
This study evaluates the roles of ProtoCL and ICL within the interaction modeling module separately. 

\subsubsection{Effect of Disentanglement Module} 
The disentanglement module clearly disentangles pathology and anatomy semantics.
Comparisons between Exp. 1, 2, and Exp. 3, 4 in TABLE~\ref{table:ablation} show a drop in grounding performance while maintaining competitive or even improved classification performance. 
This observation aligns with theoretical expectations: 
Without comprehensive information exchange in interaction modeling, the disentanglement module separates pathology and anatomy semantics, and we should only use the pathology information in downstream tasks.
Consequently, model experiences reduced grounding performance unless the separated anatomy information is re-integrated via the interaction modeling module. 

As mentioned in Sec.~\ref{sec:disentangle-module}, pathology embeddings capture the types of diseases that are essential for classification tasks, while anatomy embeddings describe the locations of diseases, which are important for grounding tasks.
We separate information about two aspects of diseases into two distinct streams by disentangling pathology and anatomy semantics, and only use the pathology outputs for downstream tasks.
In this context, the pathology information remains intact or becomes more distinct, resulting in competitive classification performance.
However, with only a part of or even no cross-stream interaction modeling, the anatomy (location) information is only partially included or not included in the outputs, leading to a decrease in grounding performance.

\begin{figure}[t]
    \centering
    \includegraphics[width=0.9\linewidth, trim=0cm 0cm 0cm 0cm]{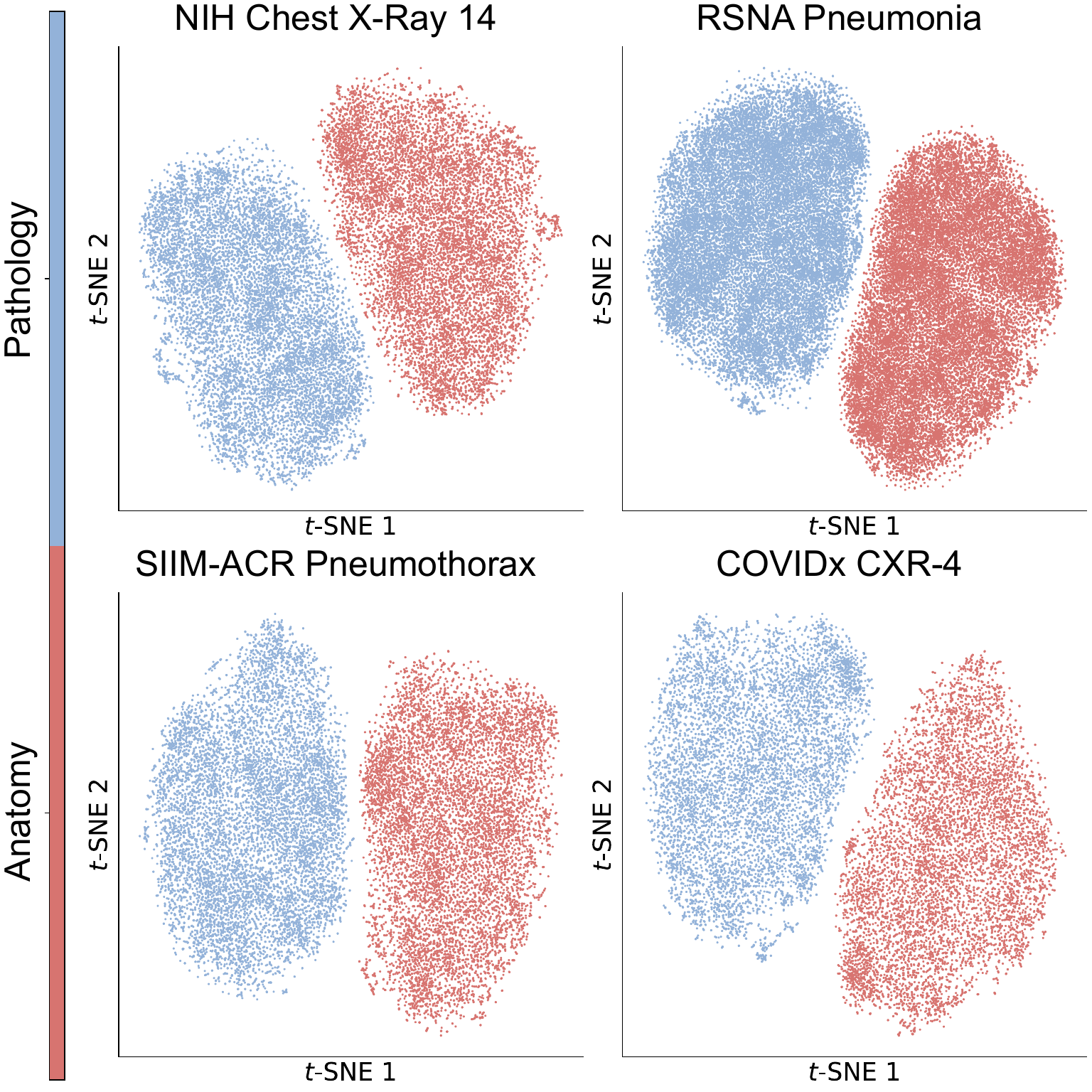}
    \caption{$t$-SNE graphs of $\ra$ and $\rp$ in evaluation datasets indicate the disentanglement module separates pathology and anatomy semantics. The red points are anatomy embeddings, and the blue ones are pathology embeddings.}
    \label{fig:PA}
\end{figure}

To further explore the quality of disentanglement, we visualize the disentangled pathology and anatomy embeddings $\ra$ and $\rp$ in $t$-SNE space, as shown in Fig.~\ref{fig:PA}. 
The visualization shows clear clusters for pathology and anatomy information across evaluation datasets, confirming that the module successfully disentangles these semantics into their respective streams.

\subsubsection{Effect of ProtoCL}

ProtoCL models cross-modal, cross-stream interaction and improve the data efficiency of positive samples. 
Comparing Exp. 1 with 3 in TABLE~\ref{table:ablation}, the model with ProtoCL achieves superior performance compared to conventional contrastive learning. 
This demonstrates improved data efficiency for positive samples when information is not disentangled.
In the case of disentangled information (Exp. 2 and 4), ProtoCL improves grounding performance by developing the relationship between anatomy information with pathology outputs. 
These results confirm that ProtoCL fulfills its design objectives of improving positive data efficiency and modeling cross-stream interaction properly.

\subsubsection{Effect of ICL}

ICL further facilitates cross-stream interaction modeling by regularizing disentangled visual pathology and anatomy semantics. 
Comparing Exp. 4 and 5 in TABLE~\ref{table:ablation}, \ourmodel\ with ICL demonstrates clear improvements across almost all metrics and datasets. 
Notably, the model’s grounding performance with both ICL and disentanglement (Exp. 5) outperforms that of models without these components (Exp. 1 and 3).
These results indicate that ICL successfully regularizes the interaction modeling between the disentangled pathology and anatomy information in a more organized manner, further enhancing the model’s ability to perform grounding and classification tasks effectively.


\section{Conclusion}


To address the entanglement of pathology and anatomy semantics and properly model their relationships, we propose a semantic vision-language alignment pipeline: \ourmodel, \textbf{Me}dical \textbf{D}ual-\textbf{S}tream \textbf{L}anguage-\textbf{I}mage \textbf{P}re-training.
It explicitly disentangles pathology and anatomy semantics in texts and images to enhance the model’s ability to utilize these semantics effectively and properly model their interactions. 
\ourmodel\ demonstrates superior generalizability and transferability by consistently outperforming eight other baselines in all experiments. 
With its superiority, \ourmodel\ offers a robust tool to aid in complex clinical tasks and lays the groundwork for future innovations in AI-driven healthcare.


\bibliographystyle{IEEEtran}
\bibliography{reference}

\end{document}